\documentclass[]{fairmeta}

\usepackage[utf8]{inputenc}      %
\usepackage[T1]{fontenc}         %
\usepackage{algorithm,algorithmicx}
\usepackage[noend]{algpseudocode}
\usepackage{amsmath,amssymb,amsfonts,amsthm,mathtools}
\usepackage{arydshln}            %
\usepackage{duckuments}
\usepackage{bm}
\usepackage{hyperref}            %
\usepackage{url}                 %
\usepackage{booktabs}            %
\usepackage[nice]{nicefrac}      %
\usepackage{microtype}           %
\usepackage[dvipsnames]{xcolor}  %
\usepackage{tcolorbox}           %
\usepackage{listings}            %
\usepackage{multirow}            %
\usepackage{colortbl}            %
\usepackage{cprotect}            %
\usepackage{lipsum}
\usepackage{enumitem}
\usepackage{graphicx}
\usepackage{float}
\usepackage[multiple]{footmisc}
\usepackage{caption,subcaption}
\usepackage{textcomp}
\usepackage{stfloats}
\usepackage{longtable,tabularx,booktabs,wrapfig}
\usepackage{makecell}
\usepackage{siunitx}
\usepackage{tabularray}
\tcbuselibrary{breakable,skins}
\usepackage{fancyvrb}
\usepackage{placeins}
\usepackage{wasysym}
\usepackage{dsfont}

\definecolor{darkblue}{rgb}{0, 0, 0.5}
\definecolor{cadmiumgreen}{rgb}{0.0, 0.42, 0.24}
\definecolor{cornellred}{rgb}{0.7, 0.11, 0.11}
\hypersetup{
  urlcolor=MidnightBlue,
  linkcolor=cornellred,
  citecolor=MidnightBlue,
  colorlinks=true,
}

\newcommand{\mname}{RPL}
\newcommand{\mnamef}{Retrieval Program Learning}

\usepackage{amsmath,amsfonts,bm}

\def\eqref#1{equation~\ref{#1}}

\def\1{\bm{1}}

\DeclareMathAlphabet{\mathsfit}{\encodingdefault}{\sfdefault}{m}{sl}
\SetMathAlphabet{\mathsfit}{bold}{\encodingdefault}{\sfdefault}{bx}{n}

\title{Learning to Program Retrievals: Composing Search Primitives via Reinforcement Learning}

\author[1,2,\dagger]{Kyuyoung Kim}
\author[2]{Zilin Xiao}
\author[2]{Zhiyuan Wang}
\author[2]{Avinash Atreya}
\author[2]{Ke Yang}
\author[1]{Jinwoo Shin}
\author[2]{Kecheng Hao}

\affiliation[1]{KAIST AI}
\affiliation[2]{Meta Superintelligence Labs}

\contribution[\dagger]{Work done at Meta}

\abstract{
Retrieval-augmented language models increasingly use multi-step search to solve knowledge-intensive tasks, yet existing systems typically expose retrieval through fixed pipelines or collections of atomic tools.
We introduce {\mnamef} ({\mname}), a programmable search framework that allows a policy to compose search primitives into executable programs with local state and control flow.
This enables adaptive search that can retrieve broadly, process intermediate results, and selectively construct model context without tying retrieval breadth to context consumption.
However, the resulting flexibility creates a difficult exploration problem: useful programs may require several coordinated decisions, while reinforcement learning often supplies only sparse feedback.
To address this challenge, we propose program mutation, a grammar-constrained and execution-verified procedure that generates local variants of suboptimal programs sampled during RL, converts them into training trajectories, and distills them back into the policy.
We evaluate {\mname} on open-domain question answering and corpus-level reasoning tasks, comparing it with a range of retrieval and reasoning baselines.
{\mname} consistently improves open-domain QA performance across model sizes and in- and out-of-domain benchmarks, with particularly large gains on multi-hop tasks.
Notably, a 4B model trained with {\mname} outperforms all 8B model baselines.
On corpus-level reasoning, {\mname} achieves the highest answer F1 and document-coverage F1 for both model sizes while substantially reducing context consumption and retrieval turns.
Our results show that programmable search expands the retrieval strategies that agents can learn, while program mutation provides an effective mechanism for exploring useful program structures beyond direct on-policy sampling.
}

\date{October 20, 2026}
\correspondence{Kyuyoung Kim at \email{kykim@kaist.ac.kr}}

\begin{document}

\maketitle

\section{Introduction}
\label{section:intro}

Language models increasingly use retrieval to solve knowledge-intensive tasks.
Retrieval-augmented generation grounds generation in external evidence, and search agents extend this paradigm by interleaving reasoning with multiple rounds of retrieval~\citep{lewis2020retrieval,yao2023react,trivedi2023}.
Recent methods use reinforcement learning with verifiable rewards (RLVR) to optimize this process directly from answer correctness, learning when to search, how to formulate queries, and how to use retrieved evidence~\citep{jin2025searchr,song2025r1}.
In most such systems, however, search is exposed as a single fixed action.
The system designer, not the policy, chooses the retriever, candidate breadth, reranker, and amount of evidence returned to the model.
This design simplifies learning, but requires a single retrieval pipeline to serve questions with different evidence needs.

A natural alternative is to give the policy finer-grained control over how retrieval is performed.
Recent methods expose multiple operations as tools, allowing the model, for instance, to choose between semantic and lexical search~\citep{luo2025marag,gandhi2026grasp}.
These interfaces broaden the policy's choices, but each operation remains a predefined atomic action, and more complex retrieval strategies must be expressed as sequences of tool calls.
We instead introduce {\mnamef} ({\mname}), a programmable search interface in which the policy writes short programs composed of search primitives and executes them within a sandbox.
For instance, a program can issue dense and sparse retrievals, filter intermediate candidates, and select which documents enter the model's context.
Code provides local state and control flow, allowing intermediate results to remain in program memory while only selected evidence is returned to the model.
This decouples retrieval breadth from context consumption and permits adaptive search procedures that cannot be expressed by a single retriever or a fixed pipeline.

Programmability, however, also creates a harder exploration problem.
Consider a multi-hop question for which neither dense nor sparse retrieval alone finds all required evidence.
A successful program may need to formulate complementary subqueries, retrieve broadly with both methods, fuse and deduplicate the results, and rerank them into a small context, potentially spanning multiple turns.
Such long-horizon dependencies are difficult to learn from outcome rewards, which provide little information about which intermediate choices contributed to success or failure.
The same interface that expands what the agent can express also expands the space it must learn to explore.

We address this challenge with \textit{program mutation}, a procedure for exploring useful retrieval programs through structured mutations.
Given a suboptimal program sampled during RL, a typed grammar generates nearby variants while preserving its queries.
Mutations make localized changes to retrieval structure, such as replacing a retriever, increasing retrieval breadth, combining results, or adding reranking.
We execute these variants and convert verified ones into synthetic trajectories for training.
We integrate mutation into an iterative procedure that collects recent RL rollouts, constructs verified mutations, fine-tunes briefly on the resulting trajectories, and resumes RL.
Program mutation thus provides experience with useful structures that sparse on-policy exploration may not discover, while leaving RL free to continue optimizing both program structure and query formulation.

\begin{figure}[t]
    \centering
    \begin{subfigure}[t]{0.68\linewidth}
        \centering
        \includegraphics[width=0.95\linewidth]{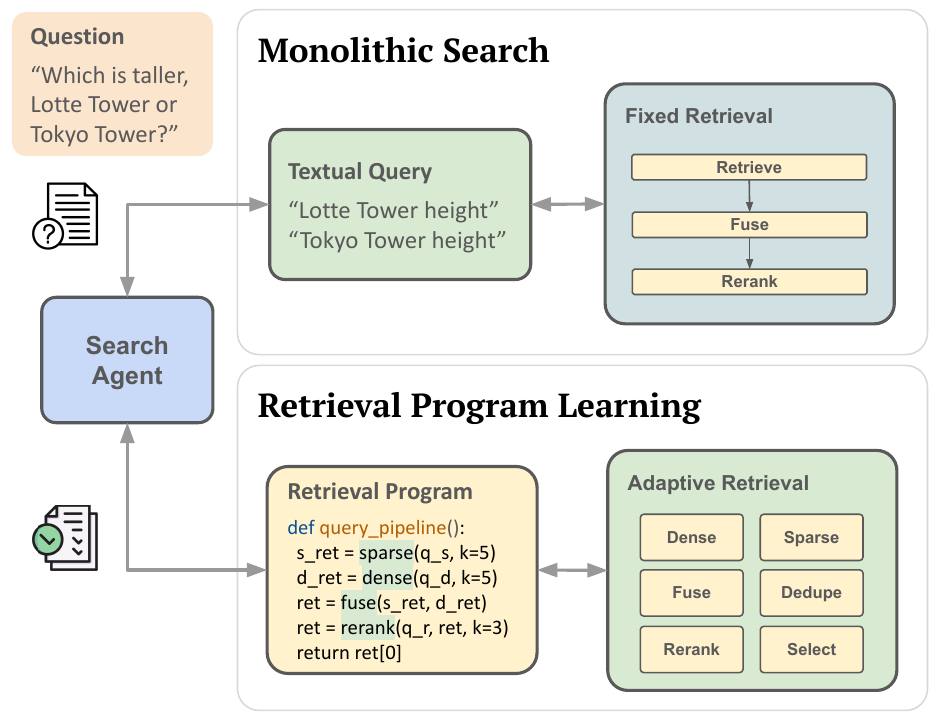}
    \end{subfigure}
    \hfill
    \begin{subfigure}[t]{0.31\linewidth}
        \centering
        \includegraphics[width=0.90\linewidth]{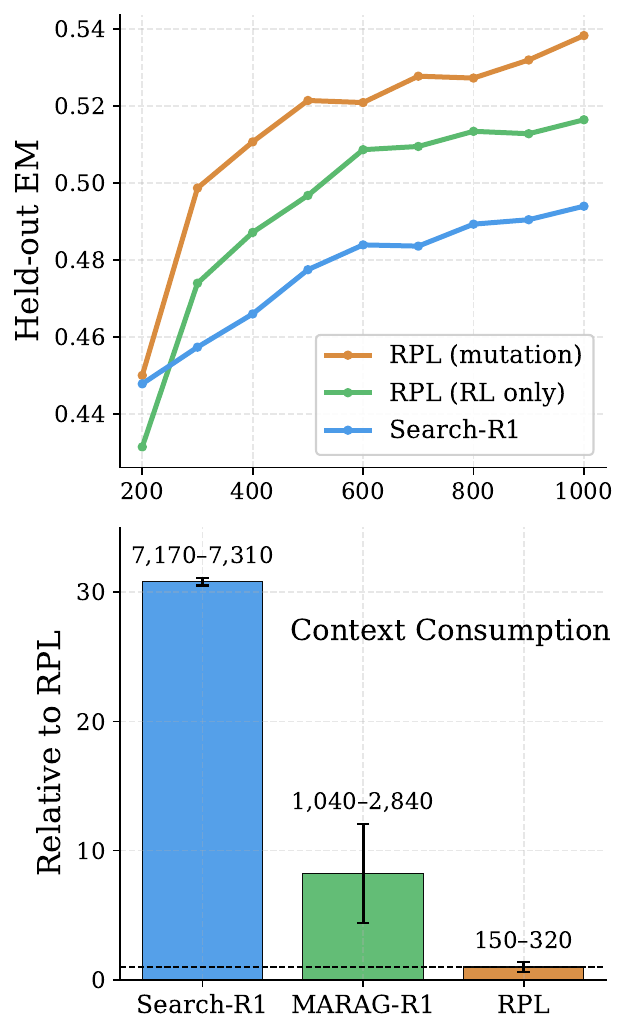}
    \end{subfigure}
    \vspace{-10pt}
    \caption{\textbf{Overview of {\mnamef}.} (1) The policy composes search primitives into executable programs rather than relying on fixed retrieval pipelines. (2) Program mutation explores local variations of suboptimal programs generated by the policy to discover more effective retrieval structures. (3) Programmable search improves answer accuracy while using context more efficiently.
    }
    \vspace{-5pt}
    \label{fig:concept}
\end{figure}

We assess programmable search in two complementary settings: open-domain question answering and corpus-level reasoning.
We train Qwen3 models on two QA datasets, Natural Questions and HotpotQA, and evaluate them across seven QA benchmarks, where {\mname} consistently outperforms retrieval-augmented RL baselines, improving average EM by up to 11.5\% for 4B and 9.0\% for 8B.
Notably, the 4B model trained with {\mname} outperforms all 8B baselines.
The gains hold across benchmarks, indicating that the learned retrieval programs generalize beyond the training tasks, with especially large gains on multi-hop tasks.
On corpus-level reasoning, {\mname} achieves the highest answer F1 and document-coverage F1 across both model sizes, improving average answer F1 by up to 58.2\% over the strongest baselines.
Moreover, {\mname}'s performance improves notably from 4B to 8B while Search-R1 and MARAG-R1 show little change, suggesting that more capable models can better leverage the additional flexibility to design effective retrieval strategies.
Further analysis shows that program mutation improves performance over RL-only training by facilitating exploration of useful structures that are often difficult to discover through on-policy RL alone.
Together, these results show that programmable search can substantially improve the quality of learned retrieval, while program mutation supports the discovery of more effective strategies.

In summary, our main contributions are as follows:
\begin{itemize}[leftmargin=7mm]
\item \textbf{Programmable search for agents.} We formulate search as programmable retrieval, where the policy composes search primitives into executable programs that adapt retrieval dynamically rather than invoking a fixed, monolithic pipeline.
\item \textbf{Exploration through program mutation.} We introduce a structured procedure for discovering enhanced retrieval programs from the policy's own suboptimal rollouts and distilling them back into the policy, augmenting on-policy exploration of program structure.
\item \textbf{Evaluation, analysis, and reproducibility.} We evaluate across diverse models and QA tasks, analyzing accuracy, context consumption, and efficiency. We release our code for training and evaluation, including sandboxed code execution and an optimized retrieval stack.
\end{itemize}

\section{Related work}
\label{section:related}

\paragraph{Retrieval-augmented generation and search agents.}

Retrieval-augmented generation grounds model outputs in external evidence to incorporate information beyond the model's parametric knowledge~\citep{lewis2020retrieval}.
ReAct~\citep{yao2023react} and IRCoT~\citep{trivedi2023} extend the standard retrieve-then-generate pipeline by interleaving reasoning with repeated tool use or retrieval.
Recent work such as Search-R1~\citep{jin2025searchr} casts search as a multi-step RL problem: the policy alternates reasoning with search, and answer correctness provides the reward for optimization.
Subsequent work improves various aspects of this framework, including evidence refinement, credit assignment, and training efficiency, while largely treating retrieval as a fixed environment action~\citep{shi2025search,zhao2025parallel,liang2026ig,chu2026redsearcher}.

Beyond implementing search as a single action, recent work has explored giving the policy finer-grained control by exposing common retrieval operations as distinct tools.
MARAG-R1~\citep{luo2025marag} and GRASP~\citep{gandhi2026grasp} train agents to coordinate retrieval operations, such as semantic and keyword search, while ProRetrieval~\citep{you2026proretrieval} leverages SQL operations for hybrid retrieval over structured and multimodal data.
Building on the broader idea of Search as Code~\citep{perplexity2026rethinking}, we take a step further by formulating search as programmable retrieval, where the policy composes search primitives into \textit{executable programs} that dynamically control the retrieval process.

\paragraph{RL exploration for language models.}

On-policy RL improves the policy using the trajectories it samples.
This makes exploration especially challenging in RL with verifiable rewards (RLVR), where sparse outcome rewards provide little guidance for intermediate decisions~\citep{kim2026correct}.
Recent analyses suggest that RLVR increases the probability of successful strategies already accessible to the base model, while strategies that are initially unlikely remain difficult to discover~\citep{yue2025does}.
To encourage broader exploration, prior work has explored prolonged optimization, entropy-aware exploration, and external guidance~\citep{liu2026prorl,dong2026rl}.

A similar exploration challenge arises in programmable search.
A richer retrieval SDK gives the policy greater flexibility, but also expands the space of possible programs.
We address this challenge with program mutation, which explores local variations of programs produced by the current policy rather than requiring useful structures to be discovered through end-to-end sampling.
This idea is related to iterative self-training in ReST~\citep{gulcehre2023reinforced}, search-based trajectory construction in Agent-R~\citep{yuan2025agent}, and execution-guided program evolution in FunSearch~\citep{romera2024mathematical}.
We specify a typed grammar that defines the set of allowable mutations, such as replacing a retriever, increasing candidate breadth, or adding reranking.
We execute trajectories with mutated programs spliced in and fine-tune on verified variants before resuming RL.
Unlike unconstrained trajectory repair, our method restricts mutations to interpretable search primitives; unlike per-instance program search, it distills the resulting variants back into the policy.

\section{Learning to program retrievals}
\label{section:method}

\subsection{Problem setup}
\label{subsec:problem_setup}

Let $x$ denote a question, $\mathcal{C}$ an external corpus, and $\mathcal{Y}(x)$ the set of reference answers.
A retrieval-augmented policy interacts with the environment over multiple turns, generating reasoning and search actions and receiving environment observations, until producing a terminal answer $\hat{y}$.
At each turn, the model can issue a search or, when it has sufficient information, terminate by producing an answer.
In the conventional setting, a search action contains a textual query $q_t$, and the environment applies a fixed retrieval function $F$, resulting in an observation $o_t$ containing external evidence: %
\begin{equation}
    o_t = F(q_t; x, \mathcal{C}).
    \label{eq:retrieval}
\end{equation}
The observation is appended to the model context before the next turn.

The policy can learn when to search and how to formulate queries, but $F$, including its retriever, candidate breadth, post-processing, and returned context, is generally fixed across questions in existing methods.
Training typically uses a verifiable outcome reward $R(\hat{y}, \mathcal{Y}(x))$, such as exact match for standard open-domain QA tasks~\citep{jin2025searchr}.
Some approaches additionally reward the quality of retrieved documents to provide denser signals for learning~\citep{luo2025marag}.

\subsection{Programmable search}

\begin{table*}[t]
    \centering
    \small %
    \caption{\textbf{Retrieval SDK.} Primitives exposed to the policy for constructing retrieval programs.}
    \label{table:sdk}
    \vspace{-5pt}
    \begin{tabular}{ll}
        \toprule[1pt]
        \textbf{Primitive} & \textbf{Description} \\
        \midrule[0.75pt]
        \verb|sparse_retrieval(q, top_k)| & Retrieve a ranked list of documents using lexical matching \\
        \verb|dense_retrieval(q, top_k)|  & Retrieve a ranked list of documents using semantic similarity \\
        \verb|fuse(*lists)|               & Fuse ranked lists using reciprocal-rank fusion \\
        \verb|dedup(docs)|                & Remove duplicates while preserving their first occurrence \\
        \verb|rerank(q, docs, top_k)|     & Rerank documents with respect to $q$ and retain the top-$k$ \\
        \bottomrule[1pt]
    \end{tabular}
    \vspace{-5pt}
\end{table*}

We introduce {\mnamef}, a programmable interface in which the action is an executable program composed of search primitives.
Rather than issuing a single textual query to a fixed retrieval function, the policy emits a program $p_t$ between \texttt{<code>} tags.
A sandbox executes the program against a retrieval SDK and returns its standard output as the observation: %
\begin{equation}
    o_t = \operatorname{ProgExec}(p_t; x, \mathcal{C}).
\end{equation}
In contrast to the fixed function $F$ in Equation~\ref{eq:retrieval}, which maps a single query $q_t$ to an observation, $p_t$ specifies the retrieval procedure itself.
A program can issue multiple retrieval calls with different queries, operate on their intermediate results, and determine which results are exposed to the model.
Intermediate results are maintained in local program state across turns, while only content explicitly printed by the program becomes $o_t$ and enters the model context.

\begin{wrapfigure}{r}{0.32\linewidth}
\vspace{-18pt}
\begin{tcolorbox}[
      colback=gray!5, 
      colframe=black!30, 
      boxrule=1.0pt,
      left=4pt, right=4pt, top=2pt, bottom=2pt,
      fonttitle=\bfseries\small,
  ]
\begin{lstlisting}[basicstyle=\ttfamily\scriptsize]
D_s = sparse_retrieval(q_s, top_k=10)
D_d = dense_retrieval(q_d, top_k=10)
D_f = fuse(dedup(D_s, D_d))
D_r = rerank(q_r, D_f, top_k=5)
print(D_r[:3])
\end{lstlisting}
\end{tcolorbox}
\vspace{-10pt}
\caption{\textbf{Example program.} A retrieval program can use multiple retrievers and selectively expose intermediate results to the model.}
\label{fig:program_example}
\vspace{-10pt}
\end{wrapfigure}

The SDK provides primitives for operations such as dense and sparse retrieval, fusion, and reranking.
Our implementation uses a small set of primitives (Table~\ref{table:sdk}), but the interface can naturally be extended with additional operations, such as metadata filtering and specialized retrievers.
The primitives operate on shared program state, allowing the policy to compose them into procedures.
For example, Figure~\ref{fig:program_example} illustrates a program that retrieves candidates using complementary retrievers with different queries, fuses and reranks their results, and exposes only three documents to the model.
Ordinary Python constructs such as variables, loops, conditionals, sorting, set operations, and arithmetic can also be used to process retrieved candidates before returning an observation.

The resulting interface decouples \textit{candidate breadth}, the number of documents retrieved, from \textit{context size}, the amount of content ultimately exposed to the model.
A program can therefore retrieve broadly, combine and process many candidates, while exposing only a small amount of most relevant evidence to the model.
For tasks that require aggregating information across many documents, this separation allows the program to process a large candidate set and return only an aggregate result, such as a list of identifiers, without exposing the underlying documents.
More broadly, this interface enables retrieval procedures that cannot be expressed by a single retriever or a fixed retrieval pipeline.

\subsection{Exploration via program mutation}
\label{subsec:program_mutation}

Programmability expands the space of retrieval strategies, but it can make effective programs harder to discover through on-policy exploration alone.
We introduce \textit{program mutation}, a structured search procedure for discovering useful retrieval structures from suboptimal programs.
Given a program $p$ sampled during RL, a typed grammar defines a set of valid mutations $\mathcal{U}(p)$, each modifying the retrieval structure while preserving the queries in the original program.
Specifically, mutations make localized changes such as replacing a retriever or increasing candidate breadth, subject to the grammar constraints.
Mutation is restricted to the final retrieval turn, since modifying an earlier turn could invalidate subsequent actions and require replaying the remainder of the trajectory.

We execute each mutated program against the same question and corpus and verify whether it retrieves sufficient evidence.
For variants that pass this check, we splice the mutated program and its execution output into the original trajectory and resample the model continuation.
All retrieval-verified variants are included in the training dataset:
\begin{equation}
\mathcal{D}_{\mathrm{mut}} =
\left\{ (x, p', y') \mid p' \in \mathcal{U}(p),\
V_{\mathrm{ret}}(p')=1 \right\},
\end{equation}
where $V_{\mathrm{ret}}$ verifies sufficient retrieved evidence and $y'$ is the resampled continuation.
We further verify the correctness of $y'$ to determine whether its answer span should be supervised or masked during training.
Thus, variants with incorrect answers can still provide supervision for their verified program structures.
Mutation provides training trajectories containing structures that the policy has not necessarily discovered through direct on-policy sampling.
See Appendix~\ref{app:program_mutation} for more details.

\subsection{Reinforcement learning for programmable search}
\label{subsec:rl_training}

\begin{algorithm}[t]
\caption{Reinforcement Learning for Programmable Search}
\label{alg:rl_program_mutation}

\begin{algorithmic}[1]

\Require Dataset $\mathcal{D}$, retrieval corpus $\mathcal{C}$, policy $\pi_\theta$,
mutation interval $N$, mutation operator $\mathcal{U}$

\For{each training step $s$}

    \State Sample batch of questions $\{x_i\}_{i=1}^B \sim \mathcal{D}$
    \State Generate trajectories $\{\tau_i \sim \pi_\theta(\cdot \mid x_i)\}_{i=1}^B$

    \State Update $\pi_\theta$ using an RL algorithm with verifiable rewards

    \If{mutation is enabled and $s \bmod N = 0$}
        \State Collect suboptimal trajectories $\mathcal{S}$ from recent rollouts
        \For{each $\tau \in \mathcal{S}$}
            \State Generate a mutation $p'$ of the last retrieval program in $\tau$
            \State Execute $p'$ to verify retrieved evidence and continuation
            \State Add verified variants to $\mathcal{D}_{\mathrm{mut}}$
        \EndFor

        \State Fine-tune $\pi_\theta$ on $\mathcal{D}_{\mathrm{mut}}$
    \EndIf
\EndFor

\State \Return $\pi_\theta$

\end{algorithmic}
\end{algorithm}

We train the policy with RL using verifiable rewards.
At each iteration, the policy generates multi-turn search trajectories and receives rewards based on the final answer and, where applicable, the quality of retrieved evidence.
The policy is then updated from these trajectories using a standard RL algorithm.
When mutation is enabled, we periodically collect suboptimal programs from recent rollouts and generate local variants.
Verified variants are converted into trajectories and used for a brief supervised fine-tuning stage.
The updated policy then resumes RL.

More concretely, the overall procedure alternates between RL training and, when enabled, program mutation and distillation:
\begin{enumerate}
    \item \textbf{Explore:} Generate search trajectories and optimize the policy with RL.
    \item \textbf{Mutate:} Collect suboptimal programs and generate verified variants.
    \item \textbf{Distill:} Fine-tune the policy on trajectories consisting of the verified programs.
\end{enumerate}

The mutation stage is not required for RL training and can be omitted when the policy already explores effective program structures.
Its role is to provide additional experience with program structures that are difficult to discover through direct on-policy sampling.
Query formulation and final policy optimization remain the responsibility of RL.

\section{Experimental results}
\label{section:experiments}

\subsection{Setup}

\paragraph{Datasets.}
We evaluate {\mname} in two complementary settings: open-domain QA and corpus-level reasoning.
For open-domain QA, we consider both single-hop datasets---Natural Questions (NQ)~\citep{kwiatkowski2019natural}, TriviaQA~\citep{joshi2017triviaqa}, and PopQA~\citep{mallen2023trust}---and multi-hop datasets---HotpotQA~\citep{yang2018hotpotqa}, 2WikiMultiHopQA (2Wiki)~\citep{ho2020constructing}, MuSiQue~\citep{trivedi2022musique}, and Bamboogle (Bamb)~\citep{press2023measuring}.
For corpus-level reasoning, we consider GlobalQA~\citep{luo2025towards}, a benchmark designed to evaluate reasoning over a document corpus.
It contains counting, sorting, top-$k$, and extremum questions that require aggregating information across documents rather than retrieving a single answer-bearing passage.
For example, some questions ask for the most recent documents with specified attributes.

\paragraph{Baselines.}
On open-domain QA, we compare against non-retrieval inference, retrieval-based inference, and training-based search methods.
For non-retrieval inference, we consider direct inference and Chain-of-Thought (CoT) reasoning.
For retrieval-based inference, we consider RAG, IRCoT~\citep{trivedi2023}, and Search-o1~\citep{li2025search}.
For training-based search, we compare against R1-style training~\citep{guo2025deepseek}, Search-R1~\citep{jin2025searchr}, and AutoRefine~\citep{shi2025search}.
See Appendix~\ref{app:additional-q25} for comparisons with additional training-based methods.
On GlobalQA, we consider MARAG-R1~\citep{luo2025marag}, a multi-tool agent with semantic search, keyword search, filtering, and aggregation that reports state-of-the-art performance on the benchmark.

\paragraph{Training and evaluation.}
In our main experiments, we use the 4B and 8B variants of Qwen3~\citep{yang2025qwen3}.
For retrieval, we use BM25~\citep{robertson1994okapi} and E5~\citep{wang2022text} as sparse and dense retrievers, respectively, with dense retrieval used by default for methods that use a single retrieval mode.
The SDK we provide for {\mname} consists of the primitives listed in Table~\ref{table:sdk}, including both retrievers, reranking with Qwen3-Reranker-4B~\citep{qwen3embedding}, reciprocal-rank fusion (RRF), and deduplication using standard Python operations.
We enable thinking mode for inference-based methods and disable it for training-based ones, choosing the mode that performs better for each setting.
See Appendix~\ref{app:think_mode} for a detailed comparison. %

For open-domain QA, we use the 2018 Wikipedia snapshot~\citep{karpukhin2020dense} as the document corpus.
Following prior work, we train on the NQ and HotpotQA training sets and evaluate on the held-out test sets of all seven QA benchmarks, reporting exact match (EM) after standard answer normalization~\citep{jin2025searchr}.
For GlobalQA, we use the corpus provided with the benchmark and evaluate both final answer and retrieval quality.
We use F1 to measure the overlap between predicted and reference answer document IDs, and document-coverage F1 to measure the coverage of relevant documents retrieved by the model.
We apply the program mutation procedure to open-domain QA tasks.
See Appendix~\ref{app:experimental_details} for additional details.

\subsection{Main results}

\paragraph{Open-domain question answering.}

\begin{table}[t]
    \centering
    \small  %
    \caption{\textbf{Evaluation across open-domain QA benchmarks.} Exact-match accuracy on single-hop and multi-hop QA benchmarks. {\mname} achieves the best overall performance, with broadly consistent gains across individual benchmarks. $^\dagger$ and $^\star$ denote in-domain and out-of-domain datasets, respectively.}
    \label{table:main_results}
    \vspace{-5pt}
    \begin{tabular}{l*{8}{c}}
    \toprule[1pt]
    \multirow{2.5}{*}{\textbf{Methods}} 
        & \multicolumn{3}{c}{\textbf{Single-Hop QA}} 
        & \multicolumn{4}{c}{\textbf{Multi-Hop QA}} 
        & \multirow{2.5}{*}{\textbf{Avg}} \\
    \cmidrule(lr){2-4} \cmidrule(lr){5-8}
        & {NQ}$^\dagger$ & {TriviaQA}$^\star$ & {PopQA}$^\star$ & {HotpotQA}$^\dagger$ & {2Wiki}$^\star$ & {MuSiQue}$^\star$ & {Bamb}$^\star$ & \\
    \midrule[0.75pt]
    \multicolumn{9}{l}{\textbf{Qwen3-4B}} \\
    Direct                  & 0.142 & 0.362 & 0.138 & 0.170 & 0.261 & 0.026 & 0.072 & 0.167 \\
    CoT                     & 0.161 & 0.417 & 0.156 & 0.186 & 0.242 & 0.036 & 0.304 & 0.215 \\
    RAG                     & 0.319 & 0.589 & 0.389 & 0.243 & 0.140 & 0.065 & 0.208 & 0.279 \\
    IRCoT                   & 0.316 & 0.574 & 0.372 & 0.292 & 0.229 & 0.133 & 0.456 & 0.339 \\
    Search-o1               & 0.285 & 0.547 & 0.354 & 0.283 & 0.226 & 0.122 & 0.400 & 0.317 \\
    R1                      & 0.243 & 0.463 & 0.184 & 0.218 & 0.291 & 0.060 & 0.288 & 0.250 \\
    AutoRefine              & 0.499 & 0.650 & 0.467 & 0.472 & 0.417 & \underline{0.217} & \underline{0.512} & 0.462 \\
    Search-R1               & \underline{0.507} & \underline{0.663} & \underline{0.475} & \underline{0.481} & \underline{0.452} & 0.209 & 0.496 & \underline{0.469} \\
    {{\mname}}              & \textbf{0.536} & \textbf{0.701} & \textbf{0.517} & \textbf{0.539} & \textbf{0.491} & \textbf{0.258} & \textbf{0.560} & \textbf{0.515} \\
    \midrule[0.75pt]
    \multicolumn{9}{l}{\textbf{Qwen3-8B}} \\
    Direct                  & 0.183 & 0.465 & 0.173 & 0.196 & 0.253 & 0.041 & 0.088 & 0.200 \\
    CoT                     & 0.200 & 0.522 & 0.189 & 0.226 & 0.278 & 0.055 & 0.376 & 0.264 \\
    RAG                     & 0.337 & 0.612 & 0.389 & 0.302 & 0.200 & 0.071 & 0.312 & 0.317 \\
    IRCoT                   & 0.287 & 0.556 & 0.335 & 0.301 & 0.227 & 0.135 & 0.400 & 0.320 \\
    Search-o1               & 0.269 & 0.543 & 0.361 & 0.294 & 0.237 & 0.126 & 0.400 & 0.318 \\
    R1                      & 0.301 & 0.569 & 0.229 & 0.263 & 0.308 & 0.083 & 0.360 & 0.302 \\
    AutoRefine              & 0.512 & 0.679 & 0.484 & \underline{0.503} & \underline{0.500} & \underline{0.234} & 0.496 & 0.487 \\
    Search-R1               & \underline{0.523} & \underline{0.680} & \underline{0.491} & \underline{0.503} & 0.486 & 0.230 & \textbf{0.544} & \underline{0.494} \\
    {{\mname}}              & \textbf{0.548} & \textbf{0.716} & \textbf{0.531} & \textbf{0.555} & \textbf{0.547} & \textbf{0.282} & \underline{0.536} & \textbf{0.531} \\
    \bottomrule[1pt] 
    \end{tabular}
    \vspace{-5pt}
\end{table}

Table~\ref{table:main_results} compares {\mname} with a range of retrieval and reasoning baselines across open-domain QA benchmarks.
{\mname} achieves the best average performance with both Qwen3-4B and Qwen3-8B, attaining 0.515 and 0.531 EM, respectively.
Among inference-based baselines, direct inference performs poorly, while CoT yields modest gains.
Qwen3-8B consistently outperforms Qwen3-4B, indicating that increased model capacity and reasoning ability provide some benefit even without external information.
Retrieval further improves performance: RAG outperforms CoT, while IRCoT and Search-o1 further improve upon RAG by interleaving multiple retrievals.
However, Search-o1's strategy of providing concise information does not yield improvements over IRCoT in this setting.
In comparison, R1 remains substantially behind retrieval-based methods, highlighting the importance of external evidence for these tasks.

{\mname} consistently outperforms the retrieval-augmented RL methods Search-R1 and AutoRefine, improving average EM by 9.8\% and 11.5\% for Qwen3-4B, and by 7.5\% and 9.0\% for Qwen3-8B, respectively.
Notably, Qwen3-4B trained with {\mname} outperforms all Qwen3-8B baselines.
The improvements are largely consistent across individual benchmarks, spanning both in-domain and out-of-domain datasets, indicating that {\mname}'s learned retrieval programs generalize across tasks.
The gains are particularly pronounced on multi-hop benchmarks such as HotpotQA, 2WikiMultiHopQA, and MuSiQue.
This suggests that executable retrieval programs are particularly effective for questions requiring multi-step evidence gathering.

\paragraph{Corpus-level reasoning.}

\begin{table}[t]
    \centering
    \small  %
    \caption{\textbf{Evaluation on GlobalQA.} F1 measures the overlap between predicted and reference answer document IDs, while dF1 measures document coverage of relevant retrieved evidence. {\mname} achieves the most competitive overall performance, with particularly strong gains on Qwen3-8B.}
    \label{table:main-results-gqa}
    \vspace{-5pt}
    \begin{tabular}{l*{10}{c}}
    \toprule[1pt]
    \multirow{2.5}{*}{\textbf{Methods}} 
        & \multicolumn{2}{c}{\textbf{TopK}} 
        & \multicolumn{2}{c}{\textbf{Count}}
        & \multicolumn{2}{c}{\textbf{Sort}}
        & \multicolumn{2}{c}{\textbf{MinMax}}
        & \multicolumn{2}{c}{\textbf{Avg}} \\
    \cmidrule(lr){2-3} \cmidrule(lr){4-5} \cmidrule(lr){6-7} \cmidrule(lr){8-9} \cmidrule(lr){10-11}
        & F1 & dF1 & F1 & dF1 & F1 & dF1 & F1 & dF1 & F1 & dF1 \\
    \midrule[0.75pt]
    \multicolumn{9}{l}{\textbf{Qwen3-4B}} \\
    RAG                    & 4.97  & 12.80 & 0.40 & 13.86 & 8.45  & 11.68 & 3.23  & 12.91 & 4.27  & 12.81 \\
    IRCoT                  & 4.30  & 12.76 & 1.11 & 13.84 & 5.07  & 13.07 & 2.49  & 12.91 & 3.24  & 13.14 \\
    Search-R1              & \underline{15.82} & 20.39 & \underline{4.19} & 20.38 & \underline{24.47} & 20.21 & \underline{11.53} & 19.35 & \underline{14.00} & 20.08 \\
    MARAG-R1               & 13.26 & \underline{23.79} & \textbf{6.98} & \underline{24.08} & 16.95 & \underline{25.45} & 11.44 & \underline{24.08} & 12.16 & \underline{24.35} \\
    {{\mname}}             & \textbf{19.49} & \textbf{29.88} & 1.86 & \textbf{29.57} & \textbf{29.14} & \textbf{31.38} & \textbf{16.92} & \textbf{29.86} & \textbf{16.85} & \textbf{30.17} \\
    \midrule[0.75pt]
    \multicolumn{9}{l}{\textbf{Qwen3-8B}} \\
    RAG                    & 4.30  & 12.80 & 1.03 & 13.86 & 7.10  & 11.68 & 4.98  & 12.91 & 4.35  & 12.81 \\
    IRCoT                  & 5.65  & 12.93 & 0.93 & 15.17 & 4.48  & 12.60 & 3.98  & 12.63 & 3.76  & 13.33 \\
    Search-R1              & \underline{15.10} & 20.09 & \underline{5.12} & 19.67 & \underline{23.93} & 19.56 & \underline{10.20} & 19.44 & \underline{13.59} & 19.69 \\
    MARAG-R1               & 14.25 & \underline{23.65} & 3.26 & \underline{24.19} & 22.36 & \underline{24.86} & 8.46 & \underline{23.96} & 12.08 & \underline{24.17} \\
    {{\mname}}             & \textbf{21.37} & \textbf{32.73} & \textbf{6.05} & \textbf{32.80} & \textbf{34.58} & \textbf{34.09} & \textbf{14.43} & \textbf{32.25} & \textbf{19.11} & \textbf{32.97} \\
    \bottomrule[1pt]
    \end{tabular}
    \vspace{-5pt}
\end{table}

Table~\ref{table:main-results-gqa} compares retrieval methods on GlobalQA, where answering questions requires aggregating information across a document corpus.
{\mname} achieves the highest answer F1 and document-coverage F1 (dF1) for both models.
The performance gap is particularly substantial for Qwen3-8B, where {\mname} markedly outperforms both Search-R1 and MARAG-R1.
In terms of answer F1, {\mname} achieves the best average performance, improving over Search-R1 and MARAG-R1 by 40.6\% and 58.2\%, respectively, with substantial gains also observed for Qwen3-4B.
Across the TopK, Sort, and MinMax tasks, {\mname} achieves the highest answer F1 on each task for both models.
Interestingly, {\mname} also benefits more from increased model scale: average answer F1 and dF1 improve from 4B to 8B, whereas Search-R1 and MARAG-R1 show little change.
This suggests that the flexibility of programmable search provides structural advantages that more capable models can better leverage to design effective retrievals.
Beyond accuracy, {\mname} enables models to retrieve and process broad candidate sets without exposing all retrieved content to the context.
We analyze this separation between retrieval breadth and context consumption in Section~\ref{sec:analysis}.

\subsection{Analysis}
\label{sec:analysis}

\paragraph{Exploration and program mutation.}

\begin{figure}[t]
    \centering
    \includegraphics[width=0.85\linewidth]{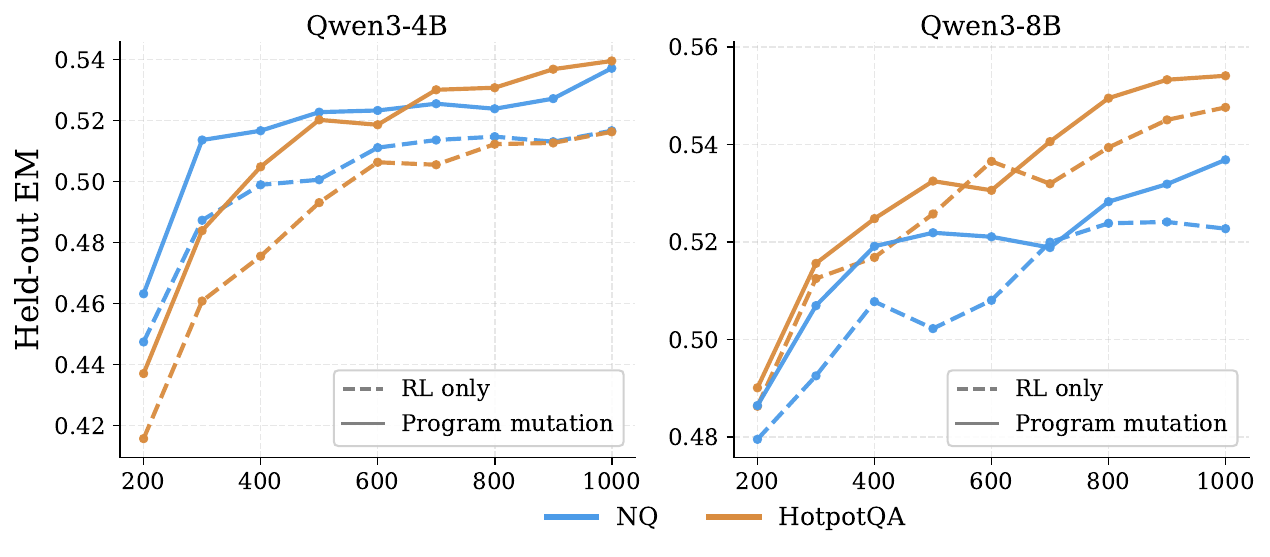}
    \vspace{-10pt}
    \caption{\textbf{Effect of program mutation on RL training.} Held-out EM for Qwen3-4B (left) and Qwen3-8B (right), with program mutation applied at step 200. Solid: RL resumed after distilling on verified program mutations. Dashed: continuous RL from the same checkpoint.}
    \label{fig:convergence}
    \vspace{-5pt}
\end{figure}

Figure~\ref{fig:convergence} shows held-out EM on NQ and HotpotQA as a function of total RL steps for both Qwen3 models, comparing RL-only training with program mutation applied once at step 200.
Distillation from the mutated programs yields a notable immediate improvement that persists throughout subsequent training.
Prior to any further RL training, the 4B policy is already ahead by 1.6 EM on NQ and 2.1 EM on HotpotQA, reflecting the benefit of supervision on the mutated programs alone.
Continued RL builds on this advantage to further widen the gap: at the final step, the 4B policy with program mutation reaches 0.536 on NQ and 0.539 on HotpotQA, compared with 0.516 and 0.517 for RL-only, corresponding to gains of 3.9\% and 4.3\%, respectively.
The average gaps over the interval are 1.7 and 2.3 EM.
The effect is similarly positive at 8B, with gains of 5.0\% and 1.5\% at the final step and average gaps of 1.0 and 0.6 EM on NQ and HotpotQA, respectively.
Although the two conditions briefly cross around steps 600--700, the mutation run again surpasses the RL-only run toward the end of training.
Notably, the gap at step 1,000 exceeds that at step 200 for both models and datasets, suggesting that subsequent RL can further build on the initial benefit from program mutation.

\paragraph{Retrieval action space and performance.}

\begin{table}[t]
    \centering
    \small  %
    \caption{\textbf{Performance and retrieval action space.} Performance under progressively richer retrieval interfaces for Qwen3-4B and Qwen3-8B. Dense: dense retrieval only; Dual: sparse and dense retrieval; SDK: full retrieval SDK; PM: program mutation applied on top of the SDK setting.}
    \label{table:sdk_ablation}
    \vspace{-5pt}
    \begin{tabular}{l*{8}{c}}
    \toprule[1pt]
    \multirow{2.5}{*}{\textbf{Methods}} 
        & \multicolumn{3}{c}{\textbf{Single-Hop QA}} 
        & \multicolumn{4}{c}{\textbf{Multi-Hop QA}} 
        & \multirow{2.5}{*}{\textbf{Avg}} \\
    \cmidrule(lr){2-4} \cmidrule(lr){5-8}
        & {NQ}$^\dagger$ & {TrivQA}$^\star$ & {PopQA}$^\star$ & {HotpotQA}$^\dagger$ & {2Wiki}$^\star$ & {MuSiQue}$^\star$ & {Bamb}$^\star$ & \\
    \midrule[0.75pt]
    \multicolumn{9}{l}{\textbf{Qwen3-4B}} \\
    {{\mname} (Dense)}       & 0.504 & 0.656 & 0.476 & 0.481 & \underline{0.475} & 0.204 & \underline{0.512} & 0.473 \\
    {{\mname} (Dual)}        & 0.502 & 0.670 & 0.487 & 0.501 & 0.471 & 0.199 & \underline{0.512} & 0.478 \\
    {{\mname} (SDK)}         & \underline{0.516} & \underline{0.684} & \underline{0.510} & \underline{0.517} & 0.463 & \underline{0.230} & 0.488 & \underline{0.487} \\
    {{\mname} (PM)}          & \textbf{0.536} & \textbf{0.701} & \textbf{0.517} & \textbf{0.539} & \textbf{0.491} & \textbf{0.258} & \textbf{0.560} & \textbf{0.515} \\
    \midrule[0.75pt]
    \multicolumn{9}{l}{\textbf{Qwen3-8B}} \\
    {{\mname} (Dense)}       & 0.514 & 0.685 & 0.486 & 0.498 & 0.446 & 0.236 & 0.528 & 0.485 \\
    {{\mname} (Dual)}        & 0.513 & 0.688 & 0.494 & 0.507 & 0.472 & 0.225 & \underline{0.536} & 0.491 \\
    {{\mname} (SDK)}         & \underline{0.522} & \underline{0.711} & \underline{0.518} & \underline{0.547} & \underline{0.521} & \textbf{0.282} & \textbf{0.552} & \underline{0.522} \\
    {{\mname} (PM)}          & \textbf{0.548} & \textbf{0.716} & \textbf{0.531} & \textbf{0.555} & \textbf{0.547} & \textbf{0.282} & \underline{0.536} & \textbf{0.531} \\
    \bottomrule[1pt]
    \end{tabular}
    \vspace{-5pt}
\end{table}

Table~\ref{table:sdk_ablation} summarizes the effect of progressively expanding the retrieval action space on open-domain QA performance.
Allowing both sparse and dense retrieval provides a small overall improvement over dense-only retrieval, with more notable gains on some multi-hop tasks.
In comparison, exposing the full retrieval SDK provides a more notable improvement: relative to the dual retriever setting, the average score increases by 1.9\% for Qwen3-4B and 6.3\% for Qwen3-8B.
Finally, program mutation further improves performance on top of the full SDK, increasing the average by another 5.7\% and 1.7\% for 4B and 8B, respectively.
This improvement is broadly consistent across benchmarks, with the only exception being a modest decrease on Bamboogle for 8B.
Overall, these results suggest that a richer action space enables the model to construct more effective retrieval pipelines, while program mutation facilitates its exploration and helps identify stronger solutions.

\paragraph{Retrieval and context efficiency.}

\begin{figure}[t]
    \centering
    \includegraphics[width=0.95\linewidth]{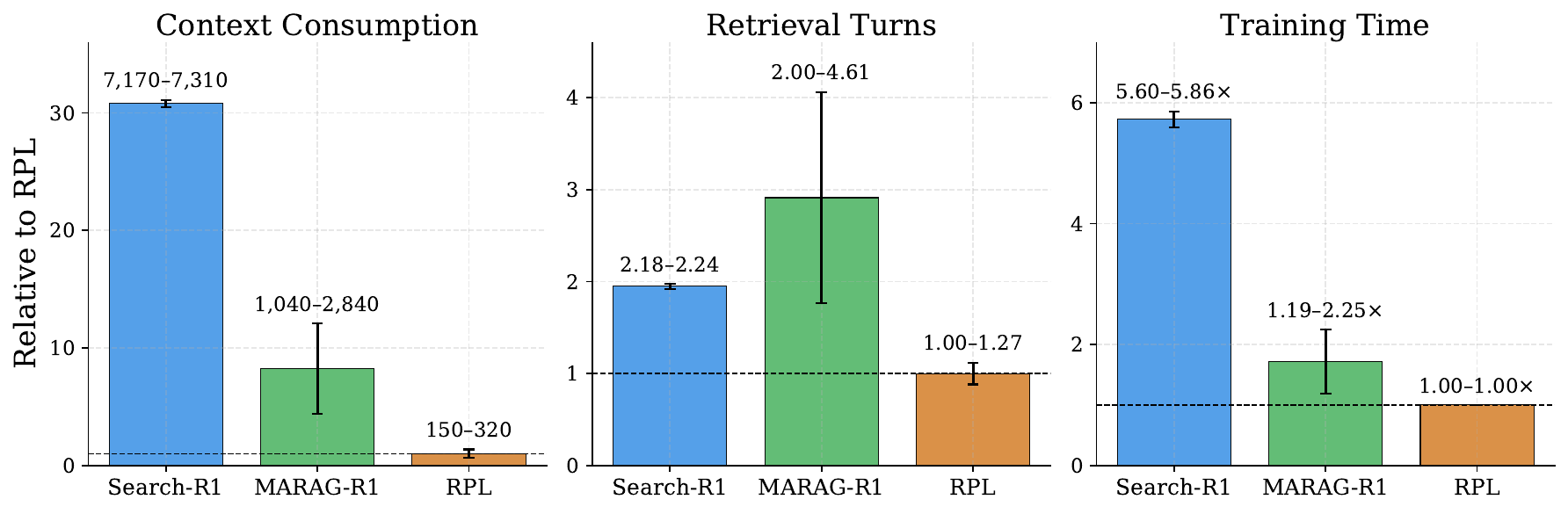}
    \vspace{-5pt}
    \caption{\textbf{Retrieval and training efficiency on GlobalQA.} Context consumption denotes the number of tokens entering the model context, retrieval turns the number of search turns, and training time the number of seconds per step. Values are normalized to the midpoint of {\mname}'s min--max range, with annotations showing the raw ranges. By programmatically controlling what enters the model context, {\mname} substantially reduces context consumption and retrieval turns.}
    \label{fig:efficiency}
    \vspace{-5pt}
\end{figure}

Beyond answer quality, programmable search provides a more efficient interface, decoupling broad retrieval from the amount of context exposed to the model.
Figure~\ref{fig:efficiency} compares {\mname} with Search-R1 and MARAG-R1 on GlobalQA in terms of context consumption, retrieval turns, and training time.
Search-R1 and MARAG-R1 expose raw documents directly to the model, coupling retrieval breadth to the amount of information placed in the context.
In contrast, {\mname} can retrieve and process a large candidate set while exposing only selected evidence to the model.
This separation allows {\mname} to achieve higher coverage with significantly less context consumption.
Programmability also reduces the interaction overhead of multi-step search.
MARAG-R1 invokes one operation per turn, requiring a separate generation and execution cycle for each tool call, whereas {\mname} can compose multiple primitives into a single program executed within one turn.
At convergence, {\mname} uses 1.00--1.27 retrieval turns, compared with 2.00--4.61 for MARAG-R1 and 2.18--2.24 for Search-R1.
Combined with reduced context, this results in substantially lower training time: {\mname} trains up to about 5.8$\times$ faster than Search-R1 and 2.2$\times$ faster than MARAG-R1.
Together, these results demonstrate the structural advantages that {\mname} provides, which decouples evidence coverage from context consumption while reducing the interaction overhead of multi-step search.

\section{Conclusion}
\label{section:conclusion}

We present {\mnamef}, a programmable search framework that allows agents to compose search primitives into executable retrieval programs.
By giving the policy control over how retrieval operations are combined, intermediate results are processed, and evidence is exposed to the model, programmable search expands the range of retrieval strategies that can be learned beyond fixed pipelines and atomic tools.
We show that this flexibility introduces a corresponding exploration challenge and propose program mutation as a mechanism for providing experience with useful program structures that are difficult to discover through on-policy RL alone.
Across open-domain question-answering and corpus-level reasoning tasks, our experiments demonstrate that programmable search can improve answer accuracy while providing explicit control over evidence coverage, context consumption, and execution resources used.

\bibliographystyle{assets/plainnat}
\bibliography{paper}

\clearpage
\newpage
\beginappendix

\section{Additional results and analyses}
\label{app:additional-results}

\subsection{Exploration across models, data, and prompts}

\begin{figure}[t]
    \centering
    \includegraphics[width=0.95\linewidth]{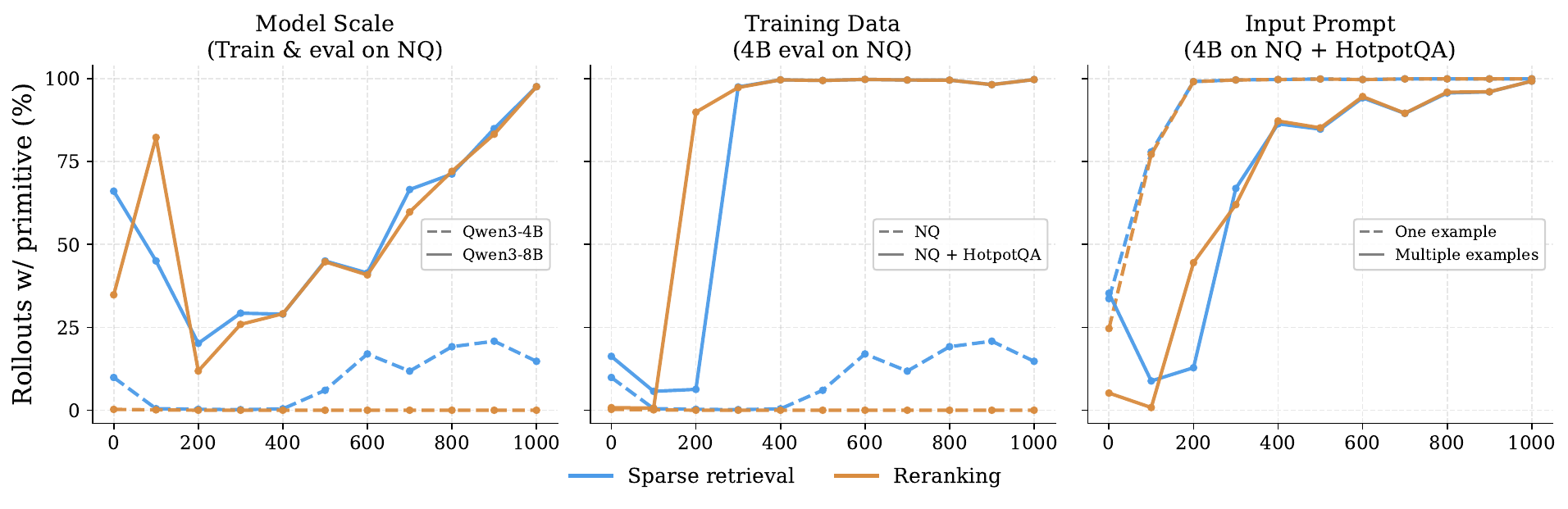}
    \vspace{-5pt}
    \caption{\textbf{Primitive adoption over RL training.} Each curve shows the percentage of held-out rollouts whose programs call sparse retrieval (blue) or reranking (orange) at least once. Panels vary one factor at a time---model scale (left), training mixture (center), and input prompt (right)---with line styles distinguishing the two conditions and the evaluation set held fixed within each.}
    \label{fig:primitive_usage}
    \vspace{-5pt}
\end{figure}

The programmable interface provides the policy with a set of search primitives, but does not require it to use any particular composition.
The retrieval strategies that emerge from RL therefore depend in part on which regions of the program space the policy explores and which behaviors are subsequently reinforced.
We analyze this by examining the programs generated during training, recording the fraction of held-out rollouts that invoke each primitive at least once.
Figure~\ref{fig:primitive_usage} varies model scale, training mixture, and prompt while holding the evaluation set fixed within each setting.
We focus on sparse retrieval and reranking, whose usage varies substantially across settings; dense retrieval, for example, is used in over 99\% of rollouts after step 200 in all settings.

\textit{Model scale} strongly affects the breadth and dynamics of exploration.
Before training, Qwen3-8B already explores substantially more programs using sparse retrieval and reranking than Qwen3-4B.
RL further amplifies this difference: reranking disappears almost immediately for the 4B model, while the 8B model initially explores, temporarily abandons, and eventually converges to programs that use both primitives in nearly all rollouts.
The smaller model stops exploring program structures that use reranking early in training, before they can be sufficiently reinforced.

The \textit{training distribution} affects which explored strategies receive sufficient reward to be reinforced.
The 4B model starts from broadly similar program distributions across the two training settings, but NQ-only training quickly converges to a dense-dominant strategy.
Adding HotpotQA instead causes both sparse retrieval and reranking to become nearly ubiquitous, even when evaluated on the same single-hop NQ questions.
The mixture thus provides opportunities to discover and reinforce program structures that are rewarded on some training examples and subsequently generalize to others.

\textit{Prompting} also affects which program structures are explored initially.
Although both prompts expose the same primitives, their demonstrated program structures lead to different dynamics.
The prompt with a single focused example quickly converges to programs that use sparse retrieval and reranking together, with their usage remaining nearly identical throughout training.
The broader prompt instead allows the two primitives to emerge more independently: reranking rises substantially before sparse retrieval catches up, and both eventually become prevalent.
The comparison suggests that broader demonstrations can expand the region of program space explored by the policy, but can also make early RL exploration less stable.

\subsection{Additional baseline comparisons}
\label{app:additional-q25}

\begin{table}[t]
    \centering
    \small  %
    \caption{\textbf{Qwen2.5 across open-domain QA benchmarks.} Exact-match accuracy on single-hop and multi-hop QA benchmarks. {\mname} achieves the best overall performance, outperforming baselines on six of the seven benchmarks. $^\dagger$ and $^\star$ indicate in-domain and out-of-domain datasets, respectively.}
    \label{table:igsearch_q25}
    \vspace{-5pt}
    \begin{tabular}{l*{8}{c}}
    \toprule[1pt]
    \multirow{2.5}{*}{\textbf{Methods}} 
        & \multicolumn{3}{c}{\textbf{Single-Hop QA}} 
        & \multicolumn{4}{c}{\textbf{Multi-Hop QA}} 
        & \multirow{2.5}{*}{\textbf{Avg}} \\
    \cmidrule(lr){2-4} \cmidrule(lr){5-8}
        & {NQ}$^\dagger$ & {TriviaQA}$^\star$ & {PopQA}$^\star$ & {HotpotQA}$^\dagger$ & {2Wiki}$^\star$ & {MuSiQue}$^\star$ & {Bamb}$^\star$ \\
    \midrule[0.75pt]
    \multicolumn{9}{l}{\textbf{Qwen2.5-3B}} \\
    ReSearch              & 0.365 & 0.571 & 0.395 & 0.351 & 0.272 & 0.095 & 0.266 & 0.331 \\
    MR-Search             & \underline{0.477} & \underline{0.635} & \underline{0.460} & 0.419 & 0.401 & 0.165 & 0.344 & 0.414 \\
    GiGPO                 & 0.420 & 0.595 & 0.424 & 0.369 & 0.370 & 0.126 & \textbf{0.641} & 0.421 \\
    IG-Search             & 0.450 & 0.614 & \underline{0.460} & \underline{0.431} & \underline{0.440} & \underline{0.191} & 0.424 & \underline{0.430} \\
    {{\mname}}            & \textbf{0.504} & \textbf{0.677} & \textbf{0.513} & \textbf{0.497} & \textbf{0.474} & \textbf{0.228} & \underline{0.432} & \textbf{0.475} \\
    \midrule[0.75pt]
    \multicolumn{9}{l}{\textbf{Qwen2.5-7B}} \\
    MR-Search             & \underline{0.502} & \underline{0.666} & 0.472 & 0.468 & 0.436 & \underline{0.221} & 0.452 & 0.460 \\
    GiGPO                 & 0.464 & 0.647 & 0.461 & 0.416 & 0.436 & 0.189 & \textbf{0.689} & 0.472 \\
    IG-Search             & 0.490 & 0.665 & \underline{0.495} & \underline{0.480} & \underline{0.445} & 0.215 & \underline{0.560} & \underline{0.479} \\
    {{\mname}}            & \textbf{0.529} & \textbf{0.700} & \textbf{0.509} & \textbf{0.541} & \textbf{0.523} & \textbf{0.269} & \underline{0.560} & \textbf{0.519} \\
    \bottomrule[1pt]
    \end{tabular}
    \vspace{-5pt}
\end{table}

To compare {\mname} with additional training-based search methods, we adopt the Qwen2.5 evaluation setting of IG-Search.
Specifically, we report the results for the existing methods, including ReSearch~\citep{chen2025research}, MR-Search~\citep{xiao2026meta}, GiGPO~\citep{feng2026group}, and IG-Search~\citep{liang2026ig}, directly from IG-Search and evaluate {\mname} with the same Qwen2.5 models.
Program mutation is not used for these experiments.

As shown in Table~\ref{table:igsearch_q25}, {\mname} achieves the highest average performance for both model sizes, reaching 0.475 and 0.519 EM, respectively.
Compared with IG-Search, the strongest prior baseline by average performance, {\mname} improves by 10.5\% and 8.4\% for Qwen2.5-3B and Qwen2.5-7B, respectively.
The improvements are broadly consistent across the individual benchmarks, with {\mname} achieving the best performance on six of seven datasets for both model sizes.
Bamboogle is the main exception, where GiGPO performs substantially better; however, the dataset contains only 125 examples, considerably fewer than the other benchmarks.
These results provide additional evidence that the benefits of programmable search extend across a broader range of language models.

\subsection{Effect of thinking mode}
\label{app:think_mode}

\begin{table*}[t]
    \centering
    \tiny  %
    \caption{\textbf{Thinking vs.\ non-thinking on open-domain QA.}
    Each cell reports non-thinking (NT) / thinking (T) performance.
    $^\dagger$ denotes a degenerate interaction between Qwen3-8B's NT mode and multi-turn agentic prompting in which the model often loops until the token budget is exhausted.}
    \label{table:think_mode}
    \vspace{-5pt}
    \setlength{\tabcolsep}{5pt}
    \begin{tabular}{l*{8}{c}}
    \toprule[1pt]
    \multirow{2.5}{*}{\textbf{Methods}}
        & \multicolumn{3}{c}{\textbf{Single-Hop QA}}
        & \multicolumn{4}{c}{\textbf{Multi-Hop QA}}
        & \multirow{2.5}{*}{\textbf{Avg}} \\
    \cmidrule(lr){2-4} \cmidrule(lr){5-8}
        & {NQ} & {TriviaQA} & {PopQA}
        & {HotpotQA} & {2Wiki} & {MuSiQue} & {Bamboogle} \\
    \midrule[0.75pt]
    \multicolumn{8}{l}{\textbf{Qwen3-4B}} \\
    CoT
        & 0.146 / 0.161 & 0.382 / 0.417 & 0.147 / 0.156
        & 0.183 / 0.186 & 0.233 / 0.242 & 0.043 / 0.036 & 0.248 / 0.304 & 0.197 / \textbf{0.215} \\
    RAG
        & 0.207 / 0.319 & 0.537 / 0.589 & 0.349 / 0.389
        & 0.221 / 0.243 & 0.149 / 0.140 & 0.047 / 0.065 & 0.176 / 0.208 & 0.241 / \textbf{0.279} \\
    IRCoT
        & 0.299 / 0.316 & 0.527 / 0.574 & 0.376 / 0.372
        & 0.256 / 0.292 & 0.211 / 0.229 & 0.090 / 0.133 & 0.264 / 0.456 & 0.289 / \textbf{0.339} \\
    Search-o1
        & 0.335 / 0.285 & 0.554 / 0.547 & 0.382 / 0.354
        & 0.270 / 0.283 & 0.197 / 0.226 & 0.084 / 0.122 & 0.232 / 0.400 & 0.293 / \textbf{0.317} \\
    Search-R1
        & 0.507 / 0.499 & 0.663 / 0.662 & 0.475 / 0.460
        & 0.481 / 0.484 & 0.452 / 0.395 & 0.209 / 0.205 & 0.496 / 0.512 & \textbf{0.469} / 0.460 \\
    \midrule[0.75pt]
    \multicolumn{8}{l}{\textbf{Qwen3-8B}} \\
    CoT
        & 0.194 / 0.200 & 0.478 / 0.522 & 0.174 / 0.189
        & 0.218 / 0.226 & 0.271 / 0.278 & 0.048 / 0.055 & 0.336 / 0.376 & 0.246 / \textbf{0.264} \\
    RAG
        & 0.342 / 0.337 & 0.611 / 0.612 & 0.401 / 0.389
        & 0.307 / 0.302 & 0.211 / 0.200 & 0.076 / 0.071 & 0.256 / 0.312 & 0.315 / \textbf{0.317} \\
    IRCoT$^\dagger$
        & 0.060 / 0.287 & 0.181 / 0.556 & 0.110 / 0.335
        & 0.121 / 0.301 & 0.060 / 0.227 & 0.039 / 0.135 & 0.072 / 0.400 & 0.092 / \textbf{0.320} \\
    Search-o1$^\dagger$
        & 0.054 / 0.269 & 0.185 / 0.543 & 0.119 / 0.361
        & 0.106 / 0.294 & 0.040 / 0.237 & 0.033 / 0.126 & 0.064 / 0.400 & 0.086 / \textbf{0.318} \\
    Search-R1
        & 0.523 / 0.505 & 0.680 / 0.682 & 0.491 / 0.484
        & 0.503 / 0.489 & 0.486 / 0.443 & 0.230 / 0.217 & 0.544 / 0.528 & \textbf{0.494} / 0.478 \\
    \bottomrule[1pt]
    \end{tabular}
    \vspace{-5pt}
\end{table*}

Table~\ref{table:think_mode} summarizes results comparing non-thinking and thinking modes on the open-domain QA benchmarks.
Thinking generally improves average performance for the inference-based methods, with the largest gains often on multi-hop tasks.
For Qwen3-4B, thinking consistently improves IRCoT on these tasks, with a particularly large gain on Bamboogle.
For Search-o1, thinking slightly reduces performance on single-hop tasks but improves performance on all multi-hop settings, again with a particularly large gain on Bamboogle.
CoT shows relatively modest changes, while RAG exhibits somewhat larger gains, particularly on single-hop tasks.
The substantial gains observed for Qwen3-8B's IRCoT and Search-o1 are largely due to a degenerate interaction between non-thinking mode and multi-turn agentic prompting: these methods occasionally emit an unrequested closing \texttt{</think>} token at the beginning of the response, leading to invalid search calls and repeated loops until the token budget is exhausted.

In contrast, Search-R1, which is trained with retrieval, shows little benefit from thinking mode: average performance is slightly lower in thinking mode for both models, with larger drops on several multi-hop tasks.
We find that training with thinking leads the model to perform fewer searches per rollout while spending substantially more tokens on reasoning, which can be particularly limiting for multi-hop questions.
Doubling the maximum reasoning length does not mitigate this issue; instead, the model simply expands its reasoning to consume the additional token budget.
Given these results, we use thinking mode for inference-based methods and non-thinking mode for training-based methods in our main evaluation.

\section{Program mutation details}
\label{app:program_mutation}

We provide further details on the program mutation procedure introduced in Section~\ref{subsec:program_mutation}.
Mutation searches for improved programs by making local, execution-verified edits to suboptimal programs produced by the current policy, enabling more effective exploration of the retrieval program space.

\subsection{Mutation grammar}
\label{subsec:pm_grammar}

\begin{figure}[t]
    \centering
    \includegraphics[width=0.95\linewidth]{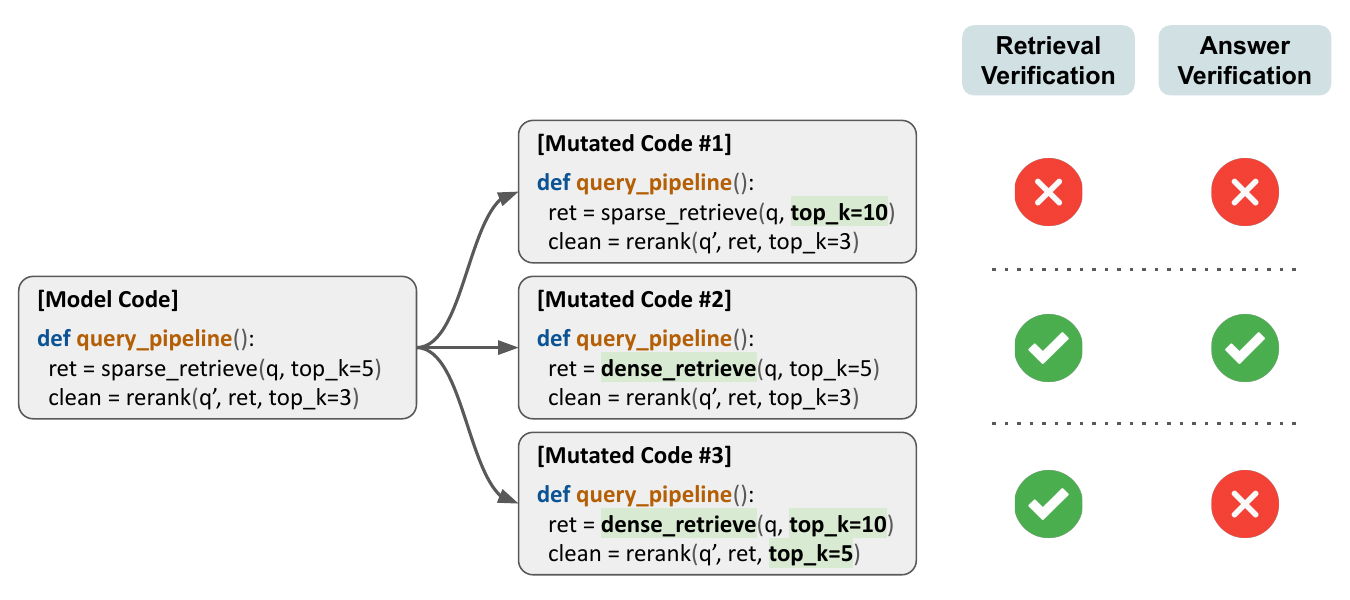}
    \vspace{-5pt}
    \caption{\textbf{Illustration of program mutation.} Given a suboptimal policy program, we generate candidates according to a mutation grammar, retaining those that satisfy the verification criteria.}
    \label{fig:mutation_example}
    \vspace{-5pt}
\end{figure}

A retrieval program can be viewed as an ordered sequence of primitives and composition operations.
We apply mutation to the program in the \emph{last retrieval turn} of a policy-generated rollout, as modifying an earlier turn could invalidate downstream policy actions.
Let $p$ denote this program and $Q_\tau$ the distinct queries appearing in $p$.
Mutation preserves the queries in $Q_\tau$ and modifies only how retrieval primitives are composed, isolating exploration to program structure while leaving RL free to optimize both structure and query formulation.
We construct candidate programs by independently choosing one value along each of four axes:
\begin{equation*}
  \underbrace{\text{query subset}}_{Q_\tau}
  \;\times\;
  \underbrace{\text{modality}}_{\{\text{dense}, \text{sparse}, \text{fuse}\}}
  \;\times\;
  \underbrace{\text{breadth}}_{\{3, 5, 10, 20, 50\}}
  \;\times\;
  \underbrace{\text{retention}}_{\{0, 3, 5, 10, 20\}}
  \label{eq:axes}
\end{equation*}
When multiple queries are selected, each query is issued with the chosen modality configuration and breadth, with reciprocal-rank fusion used to combine the resulting rankings.
A candidate may optionally apply reranking, retaining the chosen number of documents.
All calls in a candidate share a single breadth, making breadth a single decision rather than multiple independent decisions.

The mutation space is restricted to programs supported by the retrieval SDK.
A candidate must also issue at least one retrieval, fuse results when multiple calls are present, contain no duplicate calls, and print no more documents than permitted by the observation budget.
Thus, mutation operates over a constrained space of executable programs rather than arbitrary textual edits to the policy's code.

\paragraph{Candidate ordering.}
Among candidate programs, we prioritize those with smaller structural distance from the policy's program, comparing candidates lexicographically by the following criteria:
\begin{enumerate}[itemsep=1pt]
\item whether the modality configuration changed (dense, sparse, or fused)
\item whether a reranking step was added or removed
\item whether a different subset of the policy's own queries was used
\item how many rungs the breadth moved on its ladder
\item how many rungs the reranking width moved
\end{enumerate}
Earlier criteria dominate later ones.
For example, a candidate that preserves the modality and reranking structure but changes breadth by one rung is preferred to one that changes modality at the original breadth.
Ties are broken by execution resources used.

\subsection{Candidate verification and data construction}
\label{subsec:pm_verification}

Candidate programs are verified in stages before constructing the final dataset for distillation.
We first evaluate each retrieval program to determine whether it surfaces sufficient evidence to answer the question correctly.
We then verify whether the model can correctly answer the question after observing the results retrieved by the mutated program. 
For candidates that pass verification, we construct trajectories for supervised fine-tuning (Figure~\ref{fig:mutation_example}).

\paragraph{Retrieval verification.}
Each candidate is executed against the same question and corpus as the original rollout.
A retrieval verifier $V_{\mathrm{ret}}$ determines whether the mutated program surfaces sufficient evidence, using coverage of annotated supporting passages where available and otherwise checking whether the retrieved text contains an acceptable answer. 
This serves as the first filter for candidate programs, as it is model-free and substantially cheaper than resampling the model.

\paragraph{Answer verification.}
For each candidate that passes retrieval verification, we splice its actual execution output into the original trajectory and discard all subsequent policy actions.
We then resample a continuation from $\pi_\theta$:
\begin{equation*}
\hat{y} \sim \pi_\theta(\cdot \mid x, a_1, o_1, \ldots, a_{t-1}, o_{t-1}, p', \mathcal{E}(p')),
\end{equation*}
where $p'$ is the mutated program and $\mathcal{E}(p')$ is its retrieved evidence.
An answer verifier $V_{\mathrm{ans}}$ then determines whether the regenerated output $\hat{y}$ contains a correct answer.

\paragraph{Trajectory construction.}
For each mutation that passes retrieval verification, we construct a training trajectory
\begin{equation*}
\tau' = \bigl(x, a_1, o_1, \ldots, a_{t-1}, o_{t-1}, p', \mathcal{E}(p'), \hat{y}\bigr),
\end{equation*}
where $\hat{y}$ is the continuation sampled during answer verification.
Everything before the mutated retrieval turn is copied unchanged from the original rollout.
The resulting trajectory therefore contains the mutated program, its actual execution output, and a model-generated continuation.
Mutations that pass retrieval verification but fail answer verification are also retained, allowing the verified program to be supervised without the incorrect answer.
We retain whether $\hat{y}$ passed $V_{\mathrm{ans}}$ as metadata for subsequent supervision masking.

\begin{algorithm}[t]
\caption{Program Mutation}
\begin{algorithmic}[1]
\Require policy $\pi_\theta$, mutation yield threshold $\rho$

\State Collect suboptimal trajectories $\mathcal{S}$ from recent RL rollouts
\State Generate retrieval-verified mutations $\mathcal{R}$ from $\mathcal{S}$

\If{$|\mathcal{R}| / |\mathcal{S}| < \rho$}
    \State Skip mutation and continue RL
\EndIf

\State Construct verified trajectories from $\mathcal{R}$
\State Fine-tune $\pi_\theta$ on the resulting trajectories

\end{algorithmic}
\end{algorithm}

\subsection{Training with mutated programs}

We distill verified trajectories into the current policy before resuming RL.
The goal is to expose the policy to useful program structures while leaving RL free to optimize query formulation and further adapt the program to individual questions.

\paragraph{Supervision masks.}
We fine-tune the policy on the trajectories, masking the loss so that only selected spans contribute.
Injected environment observations are never supervised, while the repaired program is always supervised.
We retain every candidate that passes retrieval verification, including those whose regenerated answer fails answer verification.
For such trajectories, the answer span receives no gradient, while the repaired program remains supervised.
Thus, an incorrect regenerated answer does not prevent a trajectory from teaching the policy a retrieval program that has been independently verified to surface useful evidence.

\paragraph{RL with program mutation.}
We interleave program mutation with RL.
After every $N$ RL steps, we collect suboptimal programs from recent rollouts and generate the nearest mutation that passes retrieval verification for each program.
If the fraction of trajectories yielding such mutations falls below some threshold, we skip mutation for that round and continue RL alone.
Otherwise, we construct the verified trajectories and distill the resulting mutations before resuming RL.
In our open-domain QA experiments, a single mutation round was generally sufficient, while a second round often yielded too few candidates for training.

\section{Experimental details}
\label{app:experimental_details}

\paragraph{Training and evaluation.}
We checkpoint open-domain QA runs every 50 steps and select the one with the best average held-out accuracy on NQ and HotpotQA. For GlobalQA, we limit training to 500 steps, as some baselines take substantially longer to finish training due to the larger retrievals needed, and use the final checkpoint for evaluation.

For open-domain QA, we use exact-match accuracy for both training and evaluation. Given a set of reference answers $\mathcal{Y}(x)$, the reward is
\begin{equation*}
R_{\text{EM}} = \mathds{1}\{\hat{y} \in \mathcal{Y}(x)\},
\end{equation*}
where $\hat{y}$ is the model's predicted answer.
GlobalQA instead requires structured integer outputs based on documents satisfying specified attributes: Count asks for the number of matching documents, MinMax asks for a document ID, and Sort and TopK ask for lists of document IDs.
We parse the predicted and ground-truth answers as sets of integers and use set F1 to measure answer accuracy.
We additionally measure retrieval quality, comparing the documents retrieved throughout the rollout with the corresponding ground-truth set of relevant documents.
Let $\hat{A}$ and $A^\star$ denote the predicted and ground-truth answer sets, respectively, and let $\hat{\mathcal{D}}$ and $\mathcal{D}^\star$ denote the retrieved and gold relevant document sets, respectively.
We define
\begin{equation*}
R_{\text{ans}} = F_1(\hat{A}, A^\star),\quad
R_{\text{doc}} = F_1(\hat{\mathcal{D}}, \mathcal{D}^\star).
\end{equation*}
The training reward combines answer correctness and document coverage:
\begin{equation*}
R_{\mathrm{train}} = R_{\text{ans}} + \lambda R_{\text{doc}}.
\end{equation*}
We use set F1 rather than the token-level F1 reported by prior work because it better matches the structure of GlobalQA answers and is invariant to ordering and formatting.
For document coverage, $\hat{\mathcal{D}}$ is the union of document IDs across all retrieval calls, providing the natural analogue of the D-F1@20 metric for a policy that determines its own retrieval depth. Because the MARAG-R1 SFT dataset is unreleased, we omit its tool-exploration reward, which requires expert trajectories. Accordingly, all models in our experiments are trained with RL only.

\begin{table}[!t]
    \centering
    \small
    \caption{\textbf{Program mutation SFT hyperparameters.}}
    \label{tab:hparams-sft}
    \vspace{-5pt}
    \setlength{\tabcolsep}{8pt}
    \begin{tabular}{lll}
    \toprule[1pt]
     & \textbf{Hyperparameter} & \textbf{Value} \\
    \midrule[0.75pt]
    \textbf{Training} & Epochs & 1 \\
                      & Batch size & 32 \\
    \midrule[0.50pt]
    \textbf{Optimization} & Learning rate & 1e-6 \\
                          & LR scheduler & constant \\
                          & Warm-up steps & none \\
                          & Weight decay & 0.01 \\
                          & Gradient clipping & 1.0 \\
    \bottomrule[1pt]
    \end{tabular}
    \vspace{-5pt}
\end{table}

\begin{table}[!t]
    \centering
    \small
    \caption{\textbf{Agentic search RL hyperparameters.}}
    \label{tab:hparams-rl}
    \vspace{-5pt}
    \setlength{\tabcolsep}{8pt}
    \begin{tabular}{lll}
    \toprule[1pt]
     & \textbf{Hyperparameter} & \textbf{Value} \\
    \midrule[0.75pt]
    \textbf{Algorithm} & RL algorithm & GRPO \\
                       & Epochs per rollout & 1 \\
                       & KL coefficient & 0.001 \\
                       & Discount factor & 1.0 \\
                       & Training steps (open-domain QA) & 1,000 \\
                       & Training steps (GlobalQA) & 500 \\
                       & $R_{\text{doc}}$'s $\lambda$ (GlobalQA) & 0.5 \\
    \midrule[0.50pt]
    \textbf{Optimization} & Learning rate & 1e-6 \\
                          & LR scheduler & constant \\
                          & Warm-up steps & 150 \\
                          & Weight decay & 0.01 \\
                          & Gradient clipping & 1.0 \\
    \midrule[0.50pt]
    \textbf{Rollout} & Train batch size & 512 \\
                     & Mini-batch size & 256 \\
                     & Samples per prompt & 5 \\
                     & Temperature & 1.0 \\
    \midrule[0.50pt]
    \textbf{Search} & Max turns & 3 \\
                    & Per-turn max generation length & 2048 \\
                    & Per-turn max observation length & 2048 \\
    \bottomrule[1pt]
    \end{tabular}
    \vspace{-5pt}
\end{table}

\paragraph{Hyperparameters.}
Tables~\ref{tab:hparams-sft} and~\ref{tab:hparams-rl} summarize the key hyperparameters for supervised fine-tuning on program-mutation trajectories and agentic search RL training, respectively.
Note that RL training ran for 1,000 steps on open-domain QA and 500 steps on GlobalQA, with all other hyperparameters kept consistent across the two settings.

\paragraph{Resources.}
All training runs used 8$\times$H100 GPUs.
When reranking is required during training, we used a separate 8$\times$H100 node to serve reranking requests.
The reranker server uses multi-GPU inference, request batching, and result caching to improve throughput and efficiency.

\section{Prompt templates}

\subsection{Open-domain QA}
\begin{tcolorbox}[title={\textbf{{\mname} open-domain QA prompt}}, colback=gray!5, colframe=black!50, width=\textwidth, lowerbox=ignored, breakable] %
\begin{lstlisting}[basicstyle=\footnotesize\ttfamily, breaklines, breakindent=0pt]
Answer the given question. You must conduct reasoning inside <think> and </think> first every time you get new information. After reasoning, if you find you lack some knowledge, you can write Python between <code> and </code> to search an external knowledge corpus. The following functions are available (do not import any modules or define or call any other function):
- dense_retrieval(query: str, top_k: int): semantic search; best for paraphrased or conceptual questions (for example "what is", "who is", descriptions).
- sparse_retrieval(query: str, top_k: int): keyword search; best for exact names, numbers, dates, rare terms, or quoted strings.
- fuse(*doc_lists): merge several retrieved lists into one deduplicated, re-ranked list (reciprocal rank fusion).
- dedup(docs): remove duplicate documents from a single list.
- rerank(query: str, docs, top_k: int): reorder docs by relevance to the query with a cross-encoder and keep the top_k. Use it to turn a large, noisy candidate set into a short, focused one before reading.
Each retrieval returns a list of documents, each a dict with "title" and "text". Use print(...) to display what you want to read, and the printed output will be returned between <output> and </output>. A strong strategy is to retrieve broadly for recall (a large top_k), then rerank down to the few most relevant documents so your context stays focused and the answer is easy to find.
For a conceptual question, a small dense retrieval may be enough:
<code>
for i, doc in enumerate(dense_retrieval("what causes the aurora borealis", top_k=3)):
    print(f"Doc {i+1}(Title: {doc['title']}) {doc['text']}")
</code>
For an exact name, number, or rare term, use sparse_retrieval:
<code>
for i, doc in enumerate(sparse_retrieval("Treaty of Westphalia 1648 signatories", top_k=5)):
    print(f"Doc {i+1}(Title: {doc['title']}) {doc['text']}")
</code>
When a single retriever should surface the answer but its top hits are noisy, retrieve broadly from it alone and rerank down:
<code>
docs = dense_retrieval("what is the main cause of coral bleaching", top_k=20)
for i, doc in enumerate(rerank("what is the main cause of coral bleaching", docs, top_k=3)):
    print(f"Doc {i+1}(Title: {doc['title']}) {doc['text']}")
</code>
When the answer is hard to find, retrieve a LARGE candidate set from both retrievers, then rerank to keep only the most relevant few:
<code>
candidates = fuse(dense_retrieval("who wrote the novel Beloved", top_k=20), sparse_retrieval("Beloved novel author", top_k=20))
for i, doc in enumerate(rerank("who wrote the novel Beloved", candidates, top_k=3)):
    print(f"Doc {i+1}(Title: {doc['title']}) {doc['text']}")
</code>
For a multi-part or multi-hop question, decompose it into sub-queries, retrieve each, fuse, then rerank against the full question to surface the documents that answer it:
<code>
results = []
for q in ["director of the film Inception", "birthplace of Christopher Nolan"]:
    results.append(dense_retrieval(q, top_k=20))
    results.append(sparse_retrieval(q, top_k=20))
for i, doc in enumerate(rerank("where was the director of the film Inception born", fuse(*results), top_k=5)):
    print(f"Doc {i+1}(Title: {doc['title']}) {doc['text']}")
</code>
You can write code as many times as you want. If you find no further external knowledge needed, you can directly provide the answer inside <answer> and </answer>, without detailed illustrations. For example, <answer> Seoul </answer>.
\end{lstlisting}
\end{tcolorbox}

\subsection{GlobalQA}
\begin{tcolorbox}[title={\textbf{{\mname} GlobalQA prompt}}, colback=gray!5, colframe=black!50, width=\textwidth, lowerbox=ignored, breakable] %
\begin{lstlisting}[basicstyle=\footnotesize\ttfamily, breaklines, breakindent=0pt]
Answer the given question about a document corpus. Each document has an integer "id". Many questions ask you to COUNT the documents matching a criterion, or to find the SMALLEST/LARGEST id, SORT the ids, or list the TOP-K ids among the matching documents -- so the answer is computed from the set of matching document ids. You must conduct reasoning inside <think> and </think> first, then write Python between <code> and </code> to search the corpus. The following functions are available (do not import any modules or define or call any other function):
- dense_retrieval(query: str, top_k: int = 3): semantic search; best for paraphrased or conceptual criteria.
- sparse_retrieval(query: str, top_k: int = 3): keyword search; best for exact skills, titles, or rare terms.
- fuse(*doc_lists): merge several retrieved lists into one deduplicated, re-ranked list (reciprocal rank fusion).
- dedup(docs): remove duplicate documents from a single list.
- rerank(query: str, docs: list, top_k: int): re-score documents against the query with a cross-encoder and keep only the top_k most relevant; use it to drop non-matching documents from a broad retrieval. The documents it keeps are the ones your answer is computed from, so choosing top_k too small discards matches and makes a COUNT too low; 10, 20 or 50 are reasonable starting points.
Each retrieval returns a list of documents, best match first, each a dict with "id", "title", "text", and "score". Scores are only comparable WITHIN one list -- fuse() re-scores by rank, so its numbers are much smaller than a single retriever's. Use print(...) to display what you want to read; the printed output is returned between <output> and </output>. The answer depends on exactly the set of documents that match the criterion, so the top_k you choose IS the answer set: too small and you miss matches, too large and unrelated documents inflate a COUNT and corrupt a MinMax or SORT. There is no single right value -- pick one that fits how many documents the question is likely to match, for example 10, 20, 50, or 100. When you are unsure, retrieve a moderate top_k first, print the titles to see where the results stop being relevant, then retrieve again with an adjusted top_k before answering. When the question joins several criteria with OR, retrieve each criterion with its own short query and a depth that fits how many documents THAT criterion is likely to match -- a specific skill or title often matches only a handful, so 5 or 10 -- and fuse them; a larger depth mostly adds documents that match nothing. For a COUNT question, use len:
<code>
candidates = fuse(dense_retrieval("aviation QC inspector with in-process inspection expertise", top_k=20), sparse_retrieval("aviation quality control inspector in-process inspection", top_k=20))
ids = {d["id"] for d in rerank("aviation quality control inspector with in-process inspection expertise", candidates, top_k=20)}
print("count:", len(ids))
</code>
For a SMALLEST/LARGEST question, use min or max over the ids:
<code>
docs = fuse(sparse_retrieval("Dreamweaver", top_k=5), sparse_retrieval("Revenue Cycle Advocate", top_k=10))
ids = sorted({d["id"] for d in docs})
print("smallest id:", ids[0], "largest id:", ids[-1])
</code>
For a SORT question, sort the matching ids:
<code>
candidates = fuse(dense_retrieval("healthcare recruiters with full-cycle recruiting", top_k=20), sparse_retrieval("healthcare recruiter full cycle recruiting", top_k=20))
print(sorted({d["id"] for d in rerank("healthcare recruiters with full-cycle recruiting", candidates, top_k=20)}))
</code>
For a TOP-K question (for example the 2 smallest ids), sort then slice:
<code>
docs = fuse(sparse_retrieval("mainframe programmer", top_k=5), sparse_retrieval("COBOL developer", top_k=10))
ids = sorted({d["id"] for d in docs})
print("top 2 smallest:", ids[:2])
</code>
You can write code as many times as you want. When you know the answer, give it inside <answer> and </answer> as the number or the comma-separated ids, without detailed illustrations. For example, <answer> 8 </answer> or <answer> 330, 493 </answer>.
\end{lstlisting}
\end{tcolorbox}

\end{document}